\documentclass{article}
\usepackage{spconf2,amsmath,graphicx}
\usepackage{amsmath, amsfonts, amssymb}
\usepackage{booktabs} 
\usepackage{multirow} 
\usepackage{cite} 
\usepackage{stackengine}
\newcommand\barbelow[1]{\stackunder[1.2pt]{$#1$}{\rule{.8ex}{.075ex}}}
\usepackage{array}
\usepackage{xcolor}
\newcommand{\hl}[1]{\textcolor{black}{#1}}

\title{End-to-end Multichannel Speaker-Attributed ASR: Speaker Guided Decoder and Input Feature Analysis}
\name{Can Cui$^{1, 2}$, Imran Sheikh$^2$, Mostafa Sadeghi$^1$, Emmanuel Vincent$^1$}
\address{$^1$Université de Lorraine, CNRS, Inria, LORIA, F-54000 Nancy, France\\
$^2$Vivoka, Metz, France}

\copyrightnotice{979-8-3503-0689-7/23/\$31.00~\copyright2023 IEEE}
\begin{document}
%
\maketitle
\begin{abstract}
We present an end-to-end multichannel speaker-attributed automatic speech recognition (MC-SA-ASR) system that combines a Conformer-based encoder with multi-frame cross-channel attention and a speaker-attributed Transformer-based decoder. To the best of our knowledge, this is the first model that efficiently integrates  ASR and speaker identification modules in a multichannel setting. On simulated mixtures of LibriSpeech data, our system reduces the word error rate (WER) by up to 12\% and 16\% relative compared to previously proposed single-channel and multichannel approaches, respectively. Furthermore, we investigate the impact of different input features, including multichannel magnitude and phase information, on the ASR performance. Finally, our experiments on the AMI corpus confirm the effectiveness of our system for real-world multichannel meeting transcription. 
\end{abstract}

\vspace{-5pt}
\section{Introduction}
\label{sec:intro}
Automatic processing of multi-party (a.k.a.\ multi-speaker) speech recordings, such as meetings, requires multi-speaker automatic speech recognition (ASR) and diarization systems. The main challenges faced in these scenarios include overlapping speech, reverberation caused by distant microphones, and background noise. End-to-end multi-speaker ASR and diarization systems for single-channel \cite{guo2021multi, lu2021streaming, sklyar2021streaming, kanda2021end} and multichannel recordings \cite{chang2019mimo, chang2020end, scheibler2023end, shi2022comparative} have recently emerged, demonstrating promising results on meeting transcription tasks. 

The earlier approaches for end-to-end multi-speaker ASR and diarization for single-channel recordings lacked information sharing between the ASR and the speaker verification modules \cite{lu2021streaming} and/or required a varying number of speaker encoder (or attention) modules based on the number of speakers \cite{guo2021multi, sklyar2021streaming}. The authors of \cite{kanda2021end} proposed an end-to-end single-channel speaker-number invariant Transformer-based speaker-attributed ASR (SA-ASR)  system based on serialized output training (SOT). This system shares both speech and speaker representations across the multi-speaker ASR and speaker diarization tasks. Similar to their single-channel counterparts, the ASR and speaker identification modules in end-to-end multichannel ASR and diarization approaches \cite{chang2019mimo, chang2020end, scheibler2023end} do not share speech and speaker representations. Interestingly, the approach of multichannel word-level diarization with SOT (MC-WD-SOT) in \cite{shi2022comparative} performs a fusion of ASR and speaker information. It uses multi-frame cross-channel attention (MFCCA) \cite{yu2023mfcca} in the ASR encoder to integrate information from different channels and hidden-layer embeddings from the ASR decoder to assist speaker identification. However, the ASR and the diarization modules in this approach use different types of attention for multichannel fusion, leading to an increase in the number of model parameters. Moreover, the communication of ASR and speaker information is not bidirectional, since the speaker information is not reciprocally exploited by ASR. 

Multichannel SA-ASR (MC-SA-ASR) can exploit spatial information which is generally advantageous for localizing different speakers. However, the predominant approaches in end-to-end multichannel ASR use Mel filterbank features as input and discard phase information \cite{yu2023mfcca,zhao2021unet++}. More recently, the ``all-in-one" model in \cite{shao2022multi} uses interchannel phase difference (IPD) \cite{chen2018multi,gu2020multi,zhang2021adl} and spatial features, wherein the later require video information to localize the speakers. In \cite{chang2021end}, magnitude and phase information from each channel were separately passed through linear layers and then concatenated to form the input for multichannel ASR. To our knowledge, there is currently no research comparing and discussing the impact of different input features on MC-SA-ASR.

In this paper, we target end-to-end MC-SA-ASR, and investigate the benefit of i) leveraging speaker information for ASR and ii) exploiting phase information. To do so, we propose an MC-SA-ASR system that tightly couples a Conformer-based encoder with MFCCA, a speaker encoder and a speaker-attributed Transformer-based decoder. We explore the use of phase information as input. We conduct extensive experiments on simulated mixtures of LibriSpeech data as well as on the AMI corpus. 

The rest of the paper is organized as follows. Section \ref{sec:related-works} presents the related works. In Section \ref{sec:proposed-method}, we introduce our proposed MC-SA-ASR system and the different features and feature encoding methods. Section \ref{sec:experiments} presents our experimental setup and results on simulated data, as well as on real meeting data. Finally, Section \ref{sec:conclusion} provides a conclusion.

\begin{figure*}[!htbp]
  \centering
  \includegraphics[width=0.95\linewidth]{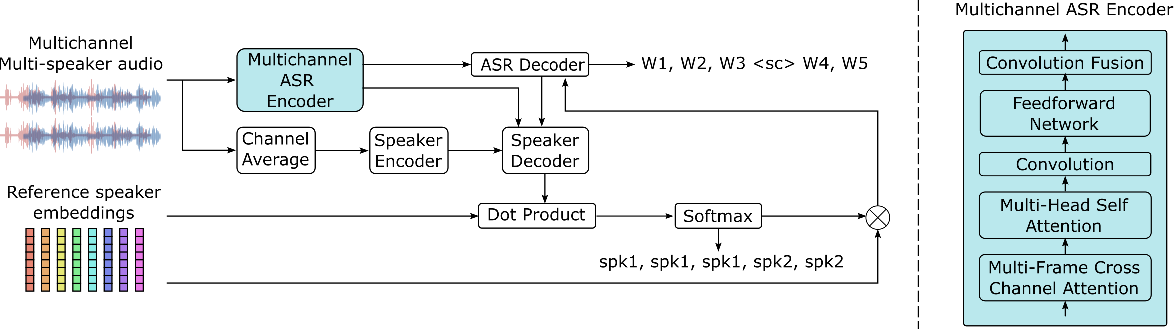}
  \vspace{-10pt}
  \caption{Overview of the proposed MC-SA-ASR system (left) and its encoder (right).}
  \label{fig:systems}
\vspace{-10pt}
\end{figure*}
\vspace{-5pt}
\section{Related works}
\label{sec:related-works}
\subsection{End-to-end single-channel SA-ASR}
\label{subsec:sa-asr}
A Transformer-based end-to-end single-channel SA-ASR (SC-SA-ASR) system was proposed in \cite{kanda2021end}. Following the SOT principle, the output is the concatenation of all speakers' sentences, where each token is associated with one speaker ID and distinct speakers are separated by a \textless sc\textgreater\ token. The inputs to the model consist of an acoustic feature sequence \(X \in \mathbb{R}^{L\times A} \) where  $L$ is the sequence length and $A$ the feature dimension, and a matrix $S \in \mathbb{R}^{E\times K}$ of $E$-dimensional reference speaker embeddings obtained from enrollment data, each corresponding to one speaker $k$ out of $K$. The feature sequence \(X\) is fed to the speaker encoder and the ASR encoder:
\begin{align}\label{eq:eq1}
  H^{\mathrm{spk}} &= \text{SpeakerEncoder}(X)\in \mathbb{R}^{T\times D}, \\
  H^{\mathrm{asr}} &= \text{ASREncoder}(X)\in \mathbb{R}^{T\times D}.
\end{align}
The resulting embeddings \(H^{\mathrm{spk}}\), \(H^{\mathrm{asr}}\) and 
the $n-1$ previous ASR output tokens
\( \hat{y}_{[1:n-1]} \) are given to the speaker decoder to obtain a speaker posterior $\hat{s}_n$ and a speaker profile \(\bar{s}_n\) associated with the $n$-th output token as follows:
\begin{align}\label{eq:eq2}
  q_n &= \text{SpeakerDecoder}(\hat{y}_{[1:n-1]},H^{\mathrm{asr}},H^{\mathrm{spk}})\in \mathbb{R}^{E},\\
  \hat{s}_n &=\text{softmax}(S^\intercal\,q_n) \in\mathbb{R}^K,\\
  \bar{s}_n & = 
  \sum{S\,\hat{s}_n}\in\mathbb{R}^{E},
\end{align}
where \((\cdot)^\intercal\) denotes matrix transposition. The ASR embedding \(H^{\mathrm{asr}}\) and the speaker profile \(\bar{s}_n\) are provided to the ASR decoder to generate the $n$-th ASR output token:
\begin{align}\label{eq:eq3}
    \hat{y}_n &= \text{ASRDecoder}(\hat{y}_{[1:n-1]},H^{\mathrm{asr}},\bar{s}_n).
\end{align}
Given the ground truth token and speaker sequences,
the training objective is to maximize the joint probability
\begin{equation}\label{eq:eq6}
\begin{aligned}
P(Y, S|X,S) &= \prod_{n=1}^{N} P(\hat{y}_n|\hat{y}_{[1:n-1]},\hat{s}_{[1:n]},X,S) \\
&\quad \times P(\hat{s}_n|\hat{y}_{[1:n-1]},\hat{s}_{[1:n-1]},X,S).
\end{aligned}
\end{equation}

\subsection{Conformer-based multichannel ASR encoder}
\label{sec:c-mc-asr}
The Conformer-based multichannel ASR encoder is based on MFCCA \cite{yu2023mfcca}, which is an attention mechanism that combines cross-channel and temporal context information. The \(h\)-th MFCCA head is computed as
\begin{align}\label{eq:eq9}
  Q_h &= XW_h^{q} +(b_h^{q})^\intercal \in \mathbb{R}^{T \times C\times D},\\
  K_h &= \barbelow{X}W_h^{k} +(b_h^{k})^\intercal \in \mathbb{R}^{T \times (2F+1)  C\times D},\\
  V_h &= \barbelow{X}W_h^{v} +(b_h^{v})^\intercal \in \mathbb{R}^{T \times (2F+1)  C\times D},\\
  H_h &= \text{softmax}\left(\frac{Q_h(K_h)^\intercal}{\sqrt{D}}\right)V_h \in \mathbb{R}^{T \times C\times D},
\end{align}
where $C$ represents the number of channels, $W_h^{q}$, $W_h^{k}$, $b_h^{q}$ and $b_h^{k}$ are learnable parameters, and \(\barbelow{X}=[\barbelow{X}^0,\ldots,\barbelow{X}^t,\ldots, \barbelow{X}^T]\) with \(\barbelow{X}^t=[X^{t-F},\ldots,X^t,\ldots,X^{t+F}]\in \mathbb{R}^{(2F+1)  C\times D} \) the concatenation of \(F\) context frames at each time step \(t\).
\vspace{-5pt}
\section{Proposed method}
\label{sec:proposed-method}
The proposed MC-SA-ASR system is illustrated in Fig.~\ref{fig:systems}. It draws inspiration from the end-to-end SC-SA-ASR model in Section \ref{subsec:sa-asr}, and uses reference speaker embeddings to guide the ASR decoder as in \eqref{eq:eq3}. However, instead of the single-channel encoder in Section \ref{subsec:sa-asr}, our system uses a modified version of the Conformer-based multichannel ASR encoder in Section \ref{sec:c-mc-asr}.  The ASR and speaker decoders remain the same as in the SC-SA-ASR model. In this section, we describe the elements that are specific to the proposed system.
\vspace{-5pt}
\subsection{Input features for multichannel ASR}
Multichannel speech signals contain spatial information that could be advantageous for discriminating speakers. Therefore, we investigate whether incorporating phase information can help our MC-SA-ASR model to generate more accurate multi-speaker transcriptions. 

\begin{figure}[t]
  \centering
  \includegraphics[width=0.99\linewidth]{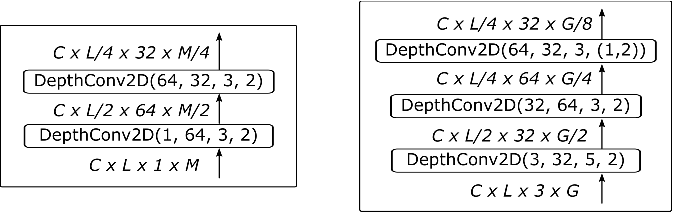}
  \caption{Depth-wise 2D-convolution feature extraction (input dimension, output dimension, kernel size, stride) for Mel filterbank (left) and magnitude+phase (right) features.}
  \label{fig:feature}
  \vspace{-4pt}
\end{figure}

We consider two alternative sets of input features. On the one hand, we compute $M$-dimensional log Mel filterbank features from the STFT magnitude only. On the other hand, we concatenate the STFT magnitude and the cosine and sine of the phase, each with $G$ dimension, into a $3 \times G$-dimensional representation, that is called the magnitude+phase feature. These features then undergo a specific processing, which is illustrated in Fig.~\ref{fig:feature} and described hereafter.

For the Mel filterbank, we apply depthwise separable convolution on each microphone. In the input array, $C$ represents the number of microphones and $L$ is the audio length. After two layers of 2-dimensional convolution, the output has a dimension of \(C \times T\times A \), where \(T=L/4 \) since each layer performs a sub-sampling of factor 2 over the time dimension, and $A=32 \times M/4$, which is the output feature dimension.

The magnitude+phase features are processed using three layers of depthwise separable convolution. In the first layer, convolutional operations are used to fuse information across magnitude, cosine and sine values. The next two layers are similar to the ones used for the Mel filterbank, resulting in features with a dimension of \(C \times T\times A \), similar to the Mel filterbank features, but with $A=32 \times G/8$. 

Finally, for both the Mel filterbank and magnitude+phase features, the output array of dimension \(C \times T\times A \) is passed to a linear layer yielding a representation array of dimension \(C \times T\times D \), where $D$ is the model dimension which is the same for both input features.
\vspace{-5pt}

\subsection{Convolution fusion}
\begin{figure}[t]
  \centering
  \includegraphics[width=0.9\linewidth]{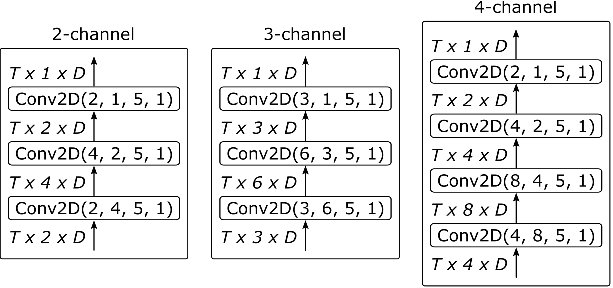}
  \vspace{-5pt}
  \caption{Multichannel convolution fusion extended to 2, 3 and 4 input channels.}
  \label{fig:conv-fusion}
  \vspace{-10pt}
\end{figure}

Convolution fusion serves as the output layer of the multichannel ASR encoder (see Fig.~\ref{fig:systems}). It combines the representations corresponding to the multiple input channels. We extend the convolution fusion from \cite{yu2023mfcca} to support 2, 3, and 4 channel input as illustrated in Fig.~\ref{fig:conv-fusion}.
\vspace{-5pt}

\subsection{Speaker encoder}
\label{sec:spk-enc}
Mel filterbank features are first
averaged across all channels 
and then fed to our speaker embedding model. We use an x-vector speaker embedding model based on the emphasized channel attention, propagation and aggregation in TDNN (ECAPA-TDNN) \cite{DBLP:conf/interspeech/DesplanquesTD20}. In order to align the dimension of the x-vectors with our model architecture, we replace the final (average pooling) layer
by a linear layer.

\vspace{-5pt}

\section{Experiments}
\label{sec:experiments}
\subsection{Evaluation on mixtures of LibriSpeech data}
\subsubsection{Dataset and metrics}
We simulate a multi-speaker scenario using the LibriSpeech dataset \cite{panayotov2015LibriSpeech}. The train-960, dev-clean and test-clean subsets
are used to generate our train, dev and test sets, respectively. We assume a linear 2- to 4-microphone array with an aperture of 10 cm. Room impulse responses (RIRs) aperture are generated by the image source method (ISM) using the gpuRIR toolkit \cite{diaz2021gpurir}. The length, width and the height of the room are randomly drawn between 3 and 8~m and between 2.4 and 3~m, respectively. The array centers and speaker positions are randomly sampled with the constraint that the array center are at most 0.5 m away from the room center, the height in the range of 0.6 to 0.8 m, and the speakers should be at least 0.5 m away from the walls. The RT60 value 
is fixed in the range [0.4, 1].
The number of speakers in each sentence is randomly chosen between 1 and 3. The speaker embeddings directory includes 8 speakers, and the speaker IDs are also randomly generated. In order to simulate multi-speaker scenarios, the second and third speakers signals are delayed relative to the previous speaker. This realistic choice also guarantees the first-in first-out (FIFO) principle behind SOT \cite{kanda2020serialized}. The sentences of each speaker are concatenated together and separated from the next speaker using the  \textless sc\textgreater\ token.

\hl{We use the word error rate (WER) to evaluate the ASR task, and the token-level speaker error rate (T-SER) and sentence-level speaker error rate (S-SER) \cite{kanda2020joint} to evaluate the speaker prediction task. In order to calculate the S-SER, a single speaker is assigned to each hypothesized sentence by taking a max over token-level speaker prediction counts within the respective sentences segmented by \textless sc\textgreater.
}
\vspace{-5pt}
\begin{table*}[!htbp]
\caption{WER (\%), sentence-level SER (S-SER) (\%) and token-level SER (T-SER) (\%) on the simulated multichannel multi-speaker LibriSpeech test set. Results are grouped by the number of speakers in the simulated mixture. The `1,2,3-speaker mix' column shows the results obtained on a test set containing mixtures of 1, 2, and 3 speakers.}
\vspace{3pt}
\centering
\label{table:test-mc-sa-asr}
\scalebox{0.75}{
\addtolength{\tabcolsep}{-0.41em}
\begin{tabular}{cccccccccccccccc}
    \toprule
    \multirow{2}{*}{\bfseries System} & 
    \multirow{2}{*}{\bfseries Input features} &
    \multirow{2}{*}{\bfseries \# Channels} & 
    \multicolumn{3}{c}{\bfseries 1-speaker  } & \multicolumn{3}{c}{\bfseries 2-speaker mix }& \multicolumn{3}{c}{\bfseries 3-speaker mix }& \multicolumn{3}{c}{\bfseries 1,2,3-speaker mix} &
    \multirow{2}{*}{\bfseries \#Param.} 
    \\ \cmidrule(lr){4-6} \cmidrule(lr){7-9} \cmidrule(lr){10-12}\cmidrule(lr){13-15}
    &&& \textbf{WER}&\textbf{S-SER} &\textbf{T-SER} & \textbf{WER}&\textbf{S-SER}&\textbf{T-SER}  & \textbf{WER}&\textbf{S-SER}&\textbf{T-SER} & \textbf{WER}&\textbf{S-SER}&\textbf{T-SER} \\ 
    \cmidrule(lr){1-16}
    \underline{Baseline Models}\\
    SC-SA-ASR \cite{kanda2021end} &Mel filterbank& 1 &  7.15 & 1.64& 3.13 & 14.68& 3.23& 5.42& 20.39& \textbf{5.59}& 7.67 &16.81 &\textbf{3.95}&6.18&61.1M\\
    \hline
    \multirow{3}{*} {MC-ASR \cite{yu2023mfcca}}&\multirow{3}{*} {Mel filterbank} & 2 & 8.52 & -& - & 16.76& -& -& 22.87& - & -&18.21 &-& -&\multirow{3}{*}{27.8M} \\
    && 3 & 8.29 & - & -& 16.54& -& -& 22.01& -& -&18.03 & -&-\\
     && 4 & 8.11 & - & -& 16.43& -& -& 21.76&-& -&17.84 & -&-\\
    \hline
    \underline{Proposed Model}\\
     &\multirow{3}{*} {Mel filterbank}
    
    & 2 & 6.64 & 1.15 &2.84 & 14.21& 3.19& 5.14& 19.03 & 5.92 &  7.34&\textbf{15.41} &4.34&6.46 &\multirow{6}{*}{51.6M} \\ 
     && 3 & \textbf{6.62} & \textbf{1.11} &\textbf{2.73} & 14.24& 3.15 & 5.56& 18.92& 6.11 &7.29 &15.25 & 4.12& 6.16\\
     && 4 & 7.17 & 1.41 & 2.95 & 14.09& \textbf{3.06}&\textbf{5.10} & 18.70& 6.06& \textbf{6.93} &14.77 &4.05&6.40\\ \cline{2-15}
     &\multirow{3}{*} {Magnitude+phase}&2 & 7.24 & 1.53& 3.30 & 15.42 & 3.85 & 6.22 & 20.34& 7.06& 7.98 &16.76 &4.87&7.29\\
     && 3 & 6.86 & 1.78& 3.39 & \textbf{13.69}& 3.72 & 5.39& 18.14& 6.86 &7.83  &\textbf{15.04} &4.08&6.28\\
    && 4 &6.91 & 1.86 & 3.43 & 13.97& 3.10&5.85 & \textbf{18.03}& 6.69& 7.26 &\textbf{14.69} &4.12&\textbf{5.85}\\
    \bottomrule
\end{tabular}
}
\vspace{-10pt}
\end{table*}
\subsubsection{Baseline}
We choose two baseline systems to study the impact of a multichannel encoder that effectively utilizes speaker information.
The first baseline is the end-to-end SC-SA-ASR system proposed in \cite{kanda2021end}. It helps us to compare the impact of a multichannel MFCCA-based ASR encoder on far-field speech. Our implementation of the SC-SA-ASR system uses ECAPA-TDNN speaker embeddings.
The second baseline is the MFCCA-based multichannel ASR (MC-ASR) model proposed in \cite{yu2023mfcca}. The MC-ASR model has been extended to perform speaker identification in the MC-WD-SOT model presented in \cite{shi2022comparative}. However, the MC-WD-SOT model does not incorporate any speaker information in the ASR decoder. Hence, the performance of the ASR module in the MC-WD-SOT model is not expected to be better than that of the MC-ASR model. Thus, in favor of a simpler implementation, we choose the MC-ASR model as the second baseline.
\vspace{-5pt}
\subsubsection{Model and training setup}
We compute STFT with a window length of 25 ms and a hop size of 10 ms. The magnitude information is used to generate 80-dimensional log Mel filterbank features. For the magnitude and phase features, we generate a $3 \times 201$-dimensional feature that includes magnitude, cosine, and sine phases. According to the convolution process in Fig.~\ref{fig:feature}, the convolutional feature extractor produces features $A$ of size 640 and 832 for the Mel filterbank features and phase-wise features, respectively. The speaker encoder uses 80-dimensional log Mel filterbank features averaged across all channels.

For all the models (SC-SA-ASR, MC-ASR and MC-SA-ASR) in our experiments, the Conformer-based encoder has 12 layers, and the Transformer-based decoder has 1 layer. The speaker decoder in SC-MC-ASR and MC-SA-ASR has 2 layers. All multi-head attention mechanisms have 4 heads,  the model dimension \(D \) is set to 256, and the size of the feedforward layer is 2048. Following \cite{yu2023mfcca}, the context frame length \(F\) of MFCCA in MC-ASR and SA-MC-ASR is set to 2. Our text tokenizer is a SentencePiece model \cite{kudo2018sentencepiece} with a vocabulary of 5000 tokens. The speaker embedding model is an ECAPA-TDNN model pre-trained on the VoxCeleb1 \cite{chung2019voxsrc} and VoxCeleb2 \cite{nagrani2020voxsrc} training data, generating 
192 dimension embeddings.
In each reference speaker embedding matrix \(S \), the number of speakers \(K\) is 
set
to 8, and the embedding for each speaker is derived from two random enrollment sentences.

Our experiments were implemented using the SpeechBrain toolkit \cite{speechbrain}. All the models were trained until convergence. The ASR modules in SC-SA-ASR and MC-SA-ASR were pre-trained for 80 epochs by setting \(S\) and \(H^{\mathrm{spk}}\) to 0. We utilized the Adam optimizer with a learning rate of $5\times 10^{-4}$ during the pre-training process. Subsequently, the ASR and speaker modules of the SC-SA-ASR model were fine-tuned for 60 epochs using a learning rate of $2.5\times 10^{-4}$. The MC-SA-ASR model was fine-tuned for 120 epochs using a learning rate of $1.5\times 10^{-5}$, after the initial training of 60 epochs. The weight of the speaker loss was set to to 0.1, following prior work \cite{kanda2020joint}. The MC-ASR model was trained for 140 epochs with a learning rate of $5\times 10^{-4}$. For all the experiments, the global batch size (batch size \(\times\)  number of GPUs \(\times\) gradient accumulation factor) is fixed to 160.
\vspace{-5pt}
\subsubsection{Results and discussion}
\label{subsubsec:results-dicsussion}
Table \ref{table:test-mc-sa-asr} presents the test results of the baseline models (SC-SA-ASR and MC-ASR) and the proposed model (MC-SA-ASR). First of all, by comparing the results of SC-SA-ASR and MC-SA-ASR, we can conclude that using a multichannel encoder to process multi-microphone speech information improves the speech recognition performance. Specifically, the MC-SA-ASR model with Mel filterbank input features achieves a WER of 14.77\% in the 4-channel scenario, that is a 12\% relative reduction compared to the SC-SA-ASR model (16.81\%). On average, the WER of the MC-SA-ASR models with 2, 3, and 4 channels (15.14\%) is reduced by 10\% relative compared to the SC-SA-ASR model. Secondly, the proposed MC-SA-ASR model, which incorporates speaker information into the ASR decoder, obtains a 16\% relative reduction in WER compared to MC-ASR (18.03\%). This suggests that leveraging speaker information can improve the performance of multi-speaker ASR in a multichannel setting.

The test WERs obtained by the proposed MC-SA-ASR model with Mel filterbank input features and with magnitude+phase input features do not exhibit significant differences. However, interestingly, in the 2-channel scenarios, Mel filterbank features (15.41\%) outperformed magnitude+phase (16.76\%) with a 8\% relative lower WER. Conversely, in the test results of the 3- and 4-channel scenarios, magnitude+phase features achieved a slight relative reduction of 1.4\% and 0.5\% in WER, respectively. This leads us to speculate that phase features may result in better ASR performance in models with a larger number of channels. However, further testing is required to validate this conclusion. Moreover, in the test results of the 3-channel scenarios, magnitude+phase features exhibited a relative reduction of 4\% in WER on both 2 and 3-speaker-mixed datasets compared to Mel filterbank features. From this, we can conclude that in multi-speaker multichannel ASR, phase features perform better on chunks with a larger number of speakers. This can be explained by the fact that as the number of speakers in a room increases, the positional information of the speakers has a greater impact on ASR performance.
\vspace{-5pt}
\subsection{Evaluation on AMI}
We validate the effectiveness of the proposed MC-SA-ASR model on real-world data by fine-tuning and testing our pre-trained model on the AMI meeting corpus \cite{carletta2005ami}.
\vspace{-5pt}
\subsubsection{Corpus preparation}
\label{sec:ami_prepare}
The AMI multiple distant microphone (MDM) corpus consists of approximately 100 hours of 8- to 16-channel audio recordings of 3 to 5 participants in meetings. The data is annotated in terms of ASR and diarization, with the start and end timestamps for each sentence. In order to process the meeting files, which typically consist of approximately 1 hour of content, we have to divide each meeting into smaller segments. We adopt a segmentation approach inspired by ``utterance groups'' \cite{kanda2021large} which works as follows. (a) Segment each meeting using a chunk size of \(b\) seconds and a hop size of \(o\) seconds. (b) If the start/end time of a segment falls within a region involving more than one speaker, it is adjusted to be two seconds outside the overlap region. (c) If the start/end time of a segment falls within a word, it is adjusted to align with the start/end of that word. The benefits of segmenting in this manner are twofold. Firstly, the utilization of a hop size allows for an increased number of training samples. Secondly, segmenting outside the speaker overlap regions respects the FIFO training approach.

\begin{figure}[t]
  \centering
  \includegraphics[width=0.95\linewidth]{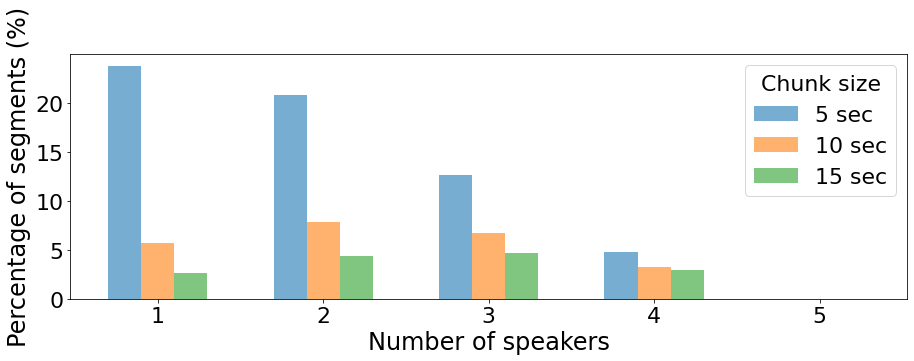}
  \vspace{-5pt}
  \caption{Percentage of segments containing a given number of speakers for different chunk sizes on the AMI corpus.}
  \label{fig:dist}
\end{figure}

We conducted experiments with chunk size of 5, 10 or 15~s, with the hop size set to 2~s. Figure~\ref{fig:dist} illustrates the distribution of the number of segments and speakers in the datasets generated using different chunk sizes. Table~\ref{table:ami} presents some statistics of the training, development, and test sets generated with a chunk size of 5~s. It is observed that the probability of a chunk containing multiple speakers increases as the chunk size increases. To evaluate the model's performance on datasets with varying numbers of speakers, we combined all the test sets segmented at 5, 10, and 15~s. We then divided them into four different test sets based on the number of speakers, which include datasets with 1, 2, 3 and 4 speakers, respectively. The number of segments for each speaker count is as follows: 5,737 segments with 1 speaker, 4,911 segments with 2 speakers, 3,986 segments with 3 speakers, and 1,936 segments with 4 speakers.
\vspace{-5pt}

\begin{table}[t]
\vspace{-5pt}
\caption{AMI statistics after segmentation (chunks size of 5 seconds). The average duration is in seconds (s), and the total duration is in hours (h).}
\label{table:ami}
\vspace{3pt}
\centering
\scalebox{0.75}{
\addtolength{\tabcolsep}{-0.41em}
\begin{tabular}{lllllll}
\hline

                  & \multicolumn{6}{c}{\bfseries Training set} \\
                  \hline
\# speakers         & 1 & 2 & 3 & 4 & 5 & total \\
\# segments    &   34719 & 30304  &18492   &6900&  10 &  90425     \\
Avg. dur. (s) &  5.62 & 6.48  &  7.74 & 8.93  &   9.43&   6.59    \\
Total dur. (h)     &  54.22 & 54.569  &  39.77 &  17.11 & 0.02  &    165.71   \\
\# words       &  444656 &  592651 &   520644&  252017 &  434 &   1810402    \\
\hline
\hline
                  & \multicolumn{6}{c}{\bfseries Development set}   \\
                  \hline
\# speakers         & 1 & 2 & 3 & 4 & & total \\
\# segments    &  4104 & 3357  &  2362 & 988  &   &  10811     \\
Avg.  dur. (s) &  5.55 &  6.38 & 7.45  &  8.45 &   &  6.49     \\
Total dur. (h)     &  6.33 &  5.95 & 4.88  & 2.31  &   &   19.50    \\
\# words       &  52730 &65072& 65575  &35803   &   &  219180     \\
\hline
\hline
                  & \multicolumn{6}{c}{\bfseries Test set}  \\
                  \hline
\# speakers         & 1 & 2 & 3 & 4 &  & total \\
\# segments    & 4211  & 3035  &   2129&  858 &   &   10233    \\
Avg. dur. (s) &  5.57 & 6.44  &  7.90 &  9.34 &   &   6.63    \\
Total dur. (h)     & 6.52  &  5.43 & 4.67  &  2.22 &   &    18.86   \\
\# words       & 52765  &   59624&  64769 &  33500 &   &   210658   \\
\hline
\end{tabular}
}
\vspace{-4pt}
\end{table}


\subsubsection{Fine-tune settings}
We fine-tune the pre-trained SC-SA-ASR and MC-SA-ASR models on the AMI MDM datasets using the Full-corpus-ASR partitions. SC-SA-ASR utilizes only the 1st channel of Array 1\footnote{Few meetings have two arrays, each array consisting of 8 microphones.} of the train-dev-test splits; while MC-SA-ASR utilizes the 1st and 5th channels for the 2-channel model.
The datasets with chunk sizes of 5, 10, and 15~s undergo fine-tuning for 40 and 90 epochs, respectively. In each case, the first half of all training epochs updates the ASR module, while the second half jointly updates the ASR and speake modules. All fine-tuning steps employ the Adam optimizer with a learning rate of $1\times 10^{-4}$, and a global batch size of 160.
\vspace{-5pt}
\begin{table}[t]
\caption{WER (\%), sentence-level SER (S-SER) (\%) and token-level SER (T-SER) (\%) of models adapted to AMI when chunk size (5, 10 or 15~s) is consistent across train, dev and test splits. Our proposed MC-SA-ASR uses 2 channels. Size represents \# of model parameters in millions.}
\label{table:mc_sa_asr_ami}
\vspace{3pt}
\centering
\scalebox{0.73}{
\addtolength{\tabcolsep}{-0.51em}
\begin{tabular}{cccccccccc}
\toprule
    \multirow{2}{*}{\bfseries System} & 
    \multicolumn{3}{c}{\bfseries 5~s} & 
    \multicolumn{3}{c}{\bfseries 10~s} & 
    \multicolumn{3}{c}{\bfseries 15~s}
    \\ \cmidrule(lr){2-4} \cmidrule(lr){5-7} \cmidrule(lr){8-10} 
    & \textbf{WER}&\textbf{S-SER}&\textbf{T-SER} & 
    \textbf{WER}&\textbf{S-SER}&\textbf{T-SER}  & 
    \textbf{WER}&\textbf{S-SER}&\textbf{T-SER}  
    \\ 
    \cmidrule(lr){1-10}
    SC-SA-ASR \cite{kanda2021end}& 46.03 &\textbf{33.86}& \textbf{28.54} & 46.54 &40.88& 28.95 & 48.78 &45.93& 30.87 \\
    \hline
    MC-SA-ASR  &\textbf{45.15}&34.14&29.08&\textbf{45.97}&\textbf{39.93}&\textbf{28.00}&\textbf{48.49}& \textbf{44.58}&\textbf{28.56}\\
\bottomrule
\end{tabular}
}
\vspace{-4pt}
\end{table}

\subsubsection{Results and discussion}
Table \ref{table:mc_sa_asr_ami}  presents the test results of SC-SA-ASR and MC-SA-ASR on datasets segmented using different chunk sizes. MC-SA-ASR consistently outperforms SC-SA-ASR across all datasets. Particularly, with a chunk size of 5~s, 2-channel MC-SA-ASR achieves a relative 2\% reduction in WER compared to SC-SA-ASR. We would also like to highlight that our proposed MC-SA-ASR model exhibits a 16\% relative reduction in model size compared to the SC-SA-ASR model which has 61M paramters. This reduction comes by replacing the first (feedforward) layer of the original Conformer by a smaller-sized MFCCA.

\begin{table}[t]
\vspace{-5pt}
\caption{WER (\%) and token-level SER (T-SER) (\%) of models adapted to AMI when chunk sizes are different across train and test splits. Rows 5~s, 10~s and 15~s refer to the chunk size in train-dev splits. The test set consists of 5, 10 and 15~s chunks. MC-SA-ASR uses 2 channels. }
\label{table:spk-ami}
\vspace{3pt}
\centering
\scalebox{0.75}{
\addtolength{\tabcolsep}{-0.41em}
\begin{tabular}{ccccccccc}
\toprule
    \multirow{2}{*}{\bfseries System} & 
    \multicolumn{2}{c}{\bfseries 1-spk} & 
    \multicolumn{2}{c}{\bfseries 2-spk mix} & 
    \multicolumn{2}{c}{\bfseries 3-spk mix}
    & \multicolumn{2}{c}{\bfseries 4-spk mix}
    \\ 
    \cmidrule(lr){2-3} \cmidrule(lr){4-5} \cmidrule(lr){6-7}
    \cmidrule(lr){8-9}
    &  \textbf{WER}&\textbf{T-SER} & \textbf{WER}&\textbf{T-SER}  & \textbf{WER}&\textbf{T-SER} 
    & \textbf{WER}&\textbf{T-SER} 
    \\ 
    \cmidrule(lr){1-9}

    SC-SA-ASR \cite{kanda2021end} \\
    \hspace{0.2cm} 5~s& 28.10 & \textbf{13.37}& 42.03 & 25.53 & 54.04 & 35.65&67.27&43.17 \\
    \hspace{0.2cm} 10~s& 31.84 & 17.25& 43.93 & 27.88 & 54.20 & 35.82 & 67.92 & 43.49 \\
    \hspace{0.2cm} 15~s& 48.38 &32.38 &60.73& 43.12 & 64.39 & 44.53& 73.68&49.40\\
    \hline
    MC-SA-ASR \\
    \hspace{0.2cm} 5~s& \textbf{27.68} & 13.43& \textbf{39.54} & \textbf{24.83} & \textbf{52.76} & 35.70&\textbf{64.72}&\textbf{41.24}\\
    \hspace{0.2cm} 10~s& 31.81 & 17.52 & 43.77 & 27.18 & 53.92 & \textbf{34.99} & 66.54 &42.80  \\

    \hspace{0.2cm} 15~s& 48.69 &31.79 &60.34& 42.48 & 63.66 & 44.19 &72.69&48.36\\

\bottomrule
\end{tabular}
}
\vspace{-4pt}
\end{table}

We also compare the performance of models trained on different chunk sizes on all the test sets. Table \ref{table:spk-ami} presents the results, divided into 4 subsets based on the number of speakers. We can observe that 2-channel MC-SA-ASR consistently achieves a lower WER than SC-SA-ASR. Particularly, the 5-second model demonstrates a 6\% relative reduction in WER compared to SC-SA-ASR on the 2-speaker test set. Moreover, we observe that models trained on smaller chunk sizes perform better. In MC-SA-ASR, the 5-second model exhibits relative reductions in WER of 43\%, 34\%, 17\%, and 11\% on the test sets with 1, 2, 3 and 4 speakers compared to the 15-second model, respectively
(note that the SER exhibits a similar trend).
This might be explained by the fact that, according to Fig.~\ref{fig:dist}, when decreasing the number of speakers, both the total number and the proportion of 5-second segments in the training set increase compared to 15-second segments.

\hl{A comparison of S-SER versus T-SER performances in Table \ref{table:mc_sa_asr_ami} shows that S-SER is much higher than T-SER on the AMI test set, for both SC-SA-ASR and MC-SA-ASR systems, as opposed to lower S-SER and higher T-SER values on the LibriSpeech test set in Table \ref{table:test-mc-sa-asr}. The reason is that the prediction of speaker change markers \textless sc\textgreater\ is a more challenging task on the AMI test set. As a further analysis, we evaluate the performance of the two systems in terms of speaker counting task on the AMI test set.}
Table \ref{table:spks-count-ami} presents the speaker counting accuracy obtained by counting the occurrences of \textless sc\textgreater\ tokens in the ASR output, on datasets with different numbers of speakers. We observe that the MC-SA-ASR system is consistently outperforming the SC-SA-ASR system.
However, the accuracy decreases by 60\% relative from the 3-speaker test set to the 4-speaker set for MC-SA-ASR (61\% for SC-SA-ASR). Furthermore, for scenarios involving 2, 3, and 4 speakers, the majority of errors originate from underestimating the number of speakers. 

\begin{table}[t]
\caption{Speaker counting accuracy (\%) on the total of 5, 10 and 15 seconds test chunks of AMI for models trained on chunks of 5 seconds. }
\vspace{3pt}
\label{table:spks-count-ami}
\centering
\scalebox{0.75}{
\addtolength{\tabcolsep}{-0.4em}
\begin{tabular}{cccccccc}
\toprule
    \multirow{2}{*}{\bfseries System} & 
    \multirow{2}{*}{\bfseries \# Speakers}& 
    \multicolumn{5}{c}{\bfseries Estimated \# speakers} & 
    \\ 
    \cmidrule(lr){3-7} 
    &&  1&2&3&4&\textgreater4
    \\ 
    \cmidrule(lr){1-7}

     \multirow{2}{*} {SC-SA-ASR \cite{kanda2021end}} & 1 & 91.75 & 7.53& 0.64& 0.06 &0.00\\
     & 2  & 11.26 & 76.82& 10.73 & 1.03 &0.14 \\
     & 3  & 1.75 & 36.67& 50.12&8.83&2.60 \\
      & 4  & 0.36 &21.02& 51.65 & 19.31&7.64\\
    \hline
    \multirow{2}{*} {MC-SA-ASR} & 1 & \textbf{92.40} & 6.98& 0.48 & 0.06&0.05 \\
     & 2  & 11.57 & \textbf{77.05}& 10.71 & 0.54&0.10\\
     & 3  & 1.80 & 37.18& \textbf{50.74}&8.50&1.75 \\
      & 4  & 0.82 & 18.59& 53.87& \textbf{20.04}&6.66 \\
\bottomrule
\end{tabular}
}
\vspace{-4pt}
\end{table}
\vspace{-5pt}
\section{Conclusion}
\label{sec:conclusion}
In this paper, we have introduced an end-to-end MC-SA-ASR system that combines a Conformer-based encoder with multi-frame cross-channel attention and a speaker-attributed Transformer-based decoder. Experimental results demonstrate that, on simulated data, our approach achieves relative reductions in WER of up to 12\% and 16\% compared to existing single-channel and multichannel methods, respectively. We also studied the impact of using Mel filterbank vs.\ magnitude+phase features on MC-SA-ASR. On real-world data, our model achieves a relative reduction in WER of up to 6\% compared to SC-SA-ASR. However, it still has limitations in accurately determining the number of speakers in scenarios involving three or more participants. Future research can focus on improving this aspect. 
\vspace{-5pt}
\section{ACKNOWLEDGMENTS}
Experiments presented in this paper were carried out using (a) the Grid'5000 testbed, supported by a scientific interest group hosted by Inria and including CNRS, RENATER and several Universities as well as other organizations (see https://www.grid5000.fr), and (b) HPC resources from GENCI-IDRIS (Grant 2023-[AD011013881]).


\clearpage

\bibliographystyle{IEEEbib}
\bibliography{mybib}

\begin{thebibliography}{10}

\bibitem{guo2021multi}
Pengcheng Guo, Xuankai Chang, Shinji Watanabe, and Lei Xie,
\newblock ``Multi-speaker {ASR} combining non-autoregressive {Conformer} {CTC}
  and conditional speaker chain,''
\newblock in {\em INTERSPEECH 2021}, 2021, pp. 1401--1405.

\bibitem{lu2021streaming}
Liang Lu, Naoyuki Kanda, Jinyu Li, and Yifan Gong,
\newblock ``Streaming multi-talker speech recognition with joint speaker
  identification,''
\newblock in {\em INTERSPEECH 2021}, 2021, pp. 1782--1782.

\bibitem{sklyar2021streaming}
Ilya Sklyar, Anna Piunova, and Yulan Liu,
\newblock ``Streaming multi-speaker {ASR} with {RNN-T},''
\newblock in {\em IEEE International Conference on Acoustics, Speech and Signal
  Processing (ICASSP)}, 2021, pp. 6903--6907.

\bibitem{kanda2021end}
Naoyuki Kanda, Guoli Ye, Yashesh Gaur, Xiaofei Wang, Zhong Meng, Zhuo Chen, and
  Takuya Yoshioka,
\newblock ``End-to-end speaker-attributed {ASR} with {Transformer},''
\newblock in {\em INTERSPEECH 2021}, 2021, pp. 4413--4417.

\bibitem{chang2019mimo}
Xuankai Chang, Wangyou Zhang, Yanmin Qian, Jonathan Le~Roux, and Shinji
  Watanabe,
\newblock ``Mimo-speech: End-to-end multi-channel multi-speaker speech
  recognition,''
\newblock in {\em 2019 IEEE Automatic Speech Recognition and Understanding
  Workshop (ASRU)}, 2019, pp. 237--244.

\bibitem{chang2020end}
Xuankai Chang, Wangyou Zhang, Yanmin Qian, Jonathan Le~Roux, and Shinji
  Watanabe,
\newblock ``End-to-end multi-speaker speech recognition with {Transformer},''
\newblock in {\em 2020 IEEE International Conference on Acoustics, Speech and
  Signal Processing (ICASSP)}, 2020, pp. 6134--6138.

\bibitem{scheibler2023end}
Robin Scheibler, Wangyou Zhang, Xuankai Chang, Shinji Watanabe, and Yanmin
  Qian,
\newblock ``End-to-end multi-speaker {ASR} with independent vector analysis,''
\newblock in {\em 2022 IEEE Spoken Language Technology Workshop (SLT)}, 2023,
  pp. 496--501.

\bibitem{shi2022comparative}
Mohan Shi, Jie Zhang, Zhihao Du, Fan Yu, Shiliang Zhang, and Li-Rong Dai,
\newblock ``A comparative study on multichannel speaker-attributed automatic
  speech recognition in multi-party meetings,''
\newblock {\em arXiv e-prints}, pp. arXiv--2211, 2022.

\bibitem{yu2023mfcca}
Fan Yu, Shiliang Zhang, Pengcheng Guo, Yuhao Liang, Zhihao Du, Yuxiao Lin, and
  Lei Xie,
\newblock ``{MFCCA}: Multi-frame cross-channel attention for multi-speaker
  {ASR} in multi-party meeting scenario,''
\newblock in {\em 2022 IEEE Spoken Language Technology Workshop (SLT)}, 2023,
  pp. 144--151.

\bibitem{zhao2021unet++}
Tuo Zhao, Yunxin Zhao, Shaojun Wang, and Mei Han,
\newblock ``Unet++-based multi-channel speech dereverberation and distant
  speech recognition,''
\newblock in {\em 12th International Symposium on Chinese Spoken Language
  Processing (ISCSLP)}, 2021, pp. 1--5.

\bibitem{shao2022multi}
Yiwen Shao, Shi-Xiong Zhang, and Dong Yu,
\newblock ``Multi-channel multi-speaker {ASR} using {3D} spatial feature,''
\newblock in {\em 2022 IEEE International Conference on Acoustics, Speech and
  Signal Processing (ICASSP)}, 2022, pp. 6067--6071.

\bibitem{chen2018multi}
Zhuo Chen, Xiong Xiao, Takuya Yoshioka, Hakan Erdogan, Jinyu Li, and Yifan
  Gong,
\newblock ``Multi-channel overlapped speech recognition with location guided
  speech extraction network,''
\newblock in {\em 2018 IEEE Spoken Language Technology Workshop (SLT)}, 2018,
  pp. 558--565.

\bibitem{gu2020multi}
Rongzhi Gu, Shi-Xiong Zhang, Yong Xu, Lianwu Chen, Yuexian Zou, and Dong Yu,
\newblock ``Multi-modal multi-channel target speech separation,''
\newblock {\em IEEE Journal of Selected Topics in Signal Processing}, vol. 14,
  no. 3, pp. 530--541, 2020.

\bibitem{zhang2021adl}
Zhuohuang Zhang, Yong Xu, Meng Yu, Shi-Xiong Zhang, Lianwu Chen, and Dong Yu,
\newblock ``{ADL-MVDR}: All deep learning {MVDR} beamformer for target speech
  separation,''
\newblock in {\em 2021 IEEE International Conference on Acoustics, Speech and
  Signal Processing (ICASSP)}, 2021, pp. 6089--6093.

\bibitem{chang2021end}
Feng-Ju Chang, Martin Radfar, Athanasios Mouchtaris, Brian King, and Siegfried
  Kunzmann,
\newblock ``End-to-end multi-channel {Transformer} for speech recognition,''
\newblock in {\em 2021 IEEE International Conference on Acoustics, Speech and
  Signal Processing (ICASSP)}, 2021, pp. 5884--5888.

\bibitem{DBLP:conf/interspeech/DesplanquesTD20}
Brecht Desplanques, Jenthe Thienpondt, and Kris Demuynck,
\newblock ``{ECAPA-TDNN:} emphasized channel attention, propagation and
  aggregation in {TDNN} based speaker verification,''
\newblock in {\em Interspeech 2020}, 2020, pp. 3830--3834.

\bibitem{panayotov2015LibriSpeech}
Vassil Panayotov, Guoguo Chen, Daniel Povey, and Sanjeev Khudanpur,
\newblock ``Librispeech: An {ASR} corpus based on public domain audio books,''
\newblock in {\em 2015 IEEE International Conference on Acoustics, Speech and
  Signal Processing (ICASSP)}, 2015, pp. 5206--5210.

\bibitem{diaz2021gpurir}
David Diaz-Guerra, Antonio Miguel, and Jose~R. Beltran,
\newblock ``{gpuRIR}: A {Python} library for room impulse response simulation
  with {GPU} acceleration,''
\newblock {\em Multimedia Tools and Applications}, vol. 80, pp. 5653--5671,
  2021.

\bibitem{kanda2020serialized}
Naoyuki Kanda, Yashesh Gaur, Xiaofei Wang, Zhong Meng, and Takuya Yoshioka,
\newblock ``Serialized output training for end-to-end overlapped speech
  recognition,''
\newblock in {\em Interspeech 2020}, 2020, pp. 2797--2801.

\bibitem{kanda2020joint}
Naoyuki Kanda, Yashesh Gaur, Xiaofei Wang, Zhong Meng, Zhuo Chen, Tianyan Zhou,
  and Takuya Yoshioka,
\newblock ``Joint speaker counting, speech recognition, and speaker
  identification for overlapped speech of any number of speakers,''
\newblock in {\em Interspeech 2020}, 2020, pp. 36--40.

\bibitem{kudo2018sentencepiece}
Taku Kudo and John Richardson,
\newblock ``Sentencepiece: A simple and language independent subword tokenizer
  and detokenizer for neural text processing,''
\newblock in {\em 2018 Conference on Empirical Methods in Natural Language
  Processing: System Demonstrations}, 2018, pp. 66--71.

\bibitem{chung2019voxsrc}
Joon~Son Chung, Arsha Nagrani, Ernesto Coto, Weidi Xie, Mitchell McLaren,
  Douglas~A. Reynolds, and Andrew Zisserman,
\newblock ``{VoxSRC} 2019: The first {VoxCeleb} speaker recognition
  challenge,''
\newblock {\em arXiv preprint arXiv:1912.02522}, 2019.

\bibitem{nagrani2020voxsrc}
Arsha Nagrani, Joon Son~Chung, Jaesung Huh, Andrew Brown, Ernesto Coto, Weidi
  Xie, Mitchell McLaren, Douglas~A. Reynolds, and Andrew Zisserman,
\newblock ``{VoxSRC} 2020: The second {VoxCeleb} speaker recognition
  challenge,''
\newblock {\em arXiv e-prints}, pp. arXiv--2012, 2020.

\bibitem{speechbrain}
Mirco Ravanelli, Titouan Parcollet, Peter Plantinga, Aku Rouhe, Samuele
  Cornell, Loren Lugosch, Cem Subakan, Nauman Dawalatabad, Abdelwahab Heba,
  Jianyuan Zhong, Ju-Chieh Chou, Sung-Lin Yeh, Szu-Wei Fu, Chien-Feng Liao,
  Elena Rastorgueva, François Grondin, William Aris, Hwidong Na, Yan Gao,
  Renato~De Mori, and Yoshua Bengio,
\newblock ``{SpeechBrain}: A general-purpose speech toolkit,'' 2021,
\newblock arXiv:2106.04624.

\bibitem{carletta2005ami}
Jean Carletta, Simone Ashby, Sebastien Bourban, Mike Flynn, Mael Guillemot,
  Thomas Hain, Jaroslav Kadlec, Vasilis Karaiskos, Wessel Kraaij, Melissa
  Kronenthal, et~al.,
\newblock ``The {AMI} meeting corpus: A pre-announcement,''
\newblock in {\em International Workshop on Machine Learning for Multimodal
  Interaction}, 2005, pp. 28--39.

\bibitem{kanda2021large}
Naoyuki Kanda, Guoli Ye, Yu~Wu, Yashesh Gaur, Xiaofei Wang, Zhong Meng, Zhuo
  Chen, and Takuya Yoshioka,
\newblock ``{Large-Scale Pre-Training of End-to-End Multi-Talker ASR for
  Meeting Transcription with Single Distant Microphone},''
\newblock in {\em Interspeech 2021}, 2021, pp. 3430--3434.

\end{thebibliography}

\end{document}